\documentclass[letterpaper, 10 pt, conference]{ieeeconf}  

\IEEEoverridecommandlockouts                              

\usepackage{amsmath,graphicx,booktabs,makecell,indentfirst,epsfig}
\usepackage{amssymb}
\usepackage{bbding}
\usepackage{multirow,multicol,color}
\usepackage{cite}
\usepackage{graphicx}
\usepackage{dsfont}
\usepackage{amsmath}
\usepackage{bbm}
\usepackage{CJK}
\usepackage{color}
\usepackage{threeparttable}
\usepackage{amssymb}
\usepackage{enumerate}
\usepackage{cite}
\usepackage{etoolbox}
\usepackage{multirow}
\usepackage[linesnumbered, ruled]{algorithm2e}
\usepackage{diagbox}

\title{\LARGE \bf
A Novel Wide-Area Multiobject Detection System with High-Probability Region Searching
}

\author{Xianlei Long$^{1}$, Hui Zhao$^{2,1}$, Chao Chen$^{1}$, Fuqiang Gu$^{1*}$, and Qingyi Gu$^{3}$
\thanks{
This work is supported by China Postdoctoral Science Foundation (No. 2023M740402), the National Natural Science Foundation of China (No. 42174050, 62322601), Venture \& Innovation Support Program for Chongqing Overseas Returnees (No. cx2021047), Chong Startup Project for Doctorate Scholars (No. CSTB2022BSXM-JSX005), the Excellent Youth Foundation of Chongqing (No. CSTB2023NSCQJQX0025), and Fundamental Research Funds for the Central Universities (No. 2023CDJXY-038, 2023CDJKYJH034).
\textit{Corresponding author: Fuqiang Gu. (e-mail: gufq@cqu.edu.cn)}}
\thanks{$^{1}$College of Computer Science, Chongqing University, Chongqing 400044, China. 
$^{2}$College of Computer Science, China University of Geoscience, Wuhan 430000, China.
$^{3}$Institute of Automation, Chinese Academy of Sciences, Beijing 100190, China.} 
}

\begin{document}

\maketitle
\thispagestyle{empty}
\pagestyle{empty}

\begin{abstract}
In recent years, wide-area visual surveillance systems have been widely applied in various industrial and transportation scenarios. These systems, however, face significant challenges when implementing multi-object detection due to conflicts arising from the need for high-resolution imaging, efficient object searching, and accurate localization. To address these challenges, this paper presents a hybrid system that incorporates a wide-angle camera, a high-speed search camera, and a galvano-mirror. In this system, the wide-angle camera offers panoramic images as prior information, which helps the search camera capture detailed images of the targeted objects. This integrated approach enhances the overall efficiency and effectiveness of wide-area visual detection systems.
Specifically, in this study, we introduce a wide-angle camera-based method to generate a panoramic probability map (PPM) for estimating high-probability regions of target object presence. Then, we propose a probability searching module that uses the PPM-generated prior information to dynamically adjust the sampling range and refine target coordinates based on uncertainty variance computed by the object detector. Finally, the integration of PPM and the probability searching module yields an efficient hybrid vision system capable of achieving 120 fps multi-object search and detection. Extensive experiments are conducted to verify the system's effectiveness and robustness.

	
\end{abstract}


\section{INTRODUCTION}

Wide-area visual detection systems, leveraging computer vision technology, have found extensive application across diverse domains, encompassing densely populated public areas~\cite{Hongbo_auto_vehicle}, airport security surveillance~\cite{Xu_pd_tracking}, sports event tracking~\cite{Yao_vt}, and intelligent transportation systems~\cite{Martin_visual_tracking}. In these varied contexts, the role of multiobject detection emerges as pivotal, enabling the recognition and interpretation of complex actions and behaviors. This capability, in turn, underpins numerous critical tasks, including surveillance monitoring~\cite{activity_monitoring, jiang_500fps_ral}, transportation detection~\cite{anomaly_detection}, et al.

However, the adoption of wide-area visual detection system faces three challenges: the pursuit of high-definition imaging for individual targets within a wide-area field of view (FOV) presents a formidable hurdle. The utilization of a wide-angle camera to cover large areas inevitably results in reduced resolution for individual objects, potentially causing the system to lose small targets. Consequently, this limitation decreases the overall accuracy of detection outcomes. (2) Existing object searching methods are not optimal when dealing with large scenes. Most of them require global random sampling of the image during the initial detection stage, which leads to low search efficiency as most sampling areas contain irrelevant backgrounds~\cite{Danielczuk_segment}. This limitation hampers real-time object detection and imposes constraints on the system's overall performance. (3) Target localization is often inaccurate in wide-area FOV due to the small size of object in an image, making accurate localization difficult. Existing image capture methods, which often use wide-angle~\cite{fisheye_2} or fish-eye lenses~\cite{fisheye_1}, exacerbate the difficulty of target localization by introducing image distortion and quality degradation~\cite{liqing_galvano}. 
To address the intricate challenges in exists wide-area visual systems, numerous methods have been developed. Some researchers have sought to attain high-definition imaging within a wide FOV by employing camera arrays comprising multiple cameras~\cite{hu2021simultaneous} or by implementing switching mirrors to simulate the functionality of multiple cameras~\cite{500_fps, Long, Three_Axis, Dual_Camera}. Nevertheless, the former approach entails intricate camera configurations, substantial costs, and manufacturing complexities. The latter approach necessitates either random scanning of the entire area or relies on wide-angle cameras for target localization, thereby introducing difficulties in achieving precise target localization.

There are many semantic probability map-based object detection paradigms~\cite{Mengers_proba_seg,Tjaden_region_tracking}, such as salient object detection (SOD) and attention mechanism-based object detection. 
SOD aims to detect the most visually conspicuous object in an image~\cite{saliency_PiCANet}. However, SOD heavily relies on high-level semantic extraction; when objects are relatively smaller in a panoramic image, the SOD often hard to extract salient features about interest objects~\cite{saliency_PAGE}, besides, the massive irrelevant background regions would hinder the high-probability information aggregation, resulting in a detection performance degradation in wide areas.
Similarly, the attention mechanism has a prominent ability to extract the most relevant region through weighted combinations, which can integrate layer-wise attention, channel-wise attention, and spatial attention to form semantic feature maps~\cite{attention_probability, attention_ASIF_Net}. Nevertheless, the probability feature map generated by the attention method can easily lose discriminating information related to the objects that are captured in a panoramic image~\cite{saliency_PiCANet}.
Despite many advanced prior probability-guided methods having been proposed, the probability maps are often coarsely computed by a CNN through the original image, which ignores the relationship between foreground objects and background regions and misses the corresponding probability of desired objects.
Different from the existing probability map generation methods, which just parse the high-probability regions. In this research, we introduce a hybrid multi-camera detection framework to generate a coarse-to-fine semantic probability map and cooperatively detect desired objects.

In recent years, uncertainty estimation has been widely explored in object searching. However, Nguyen et al.~\cite{uncertainty_pred} pointed out that uncertainty measures may not be a reliable stand-in for a reasonable uncertainty estimate. Miller et al.~\cite{uncertainty_merging} proposed an uncertainty estimation and bounding box merging strategy to refine the object coordinate. However, they ignore the relationship of objects to the environment. It's better to combine the uncertainty and the object's state to jointly optimize the search process.


Consequently, to solve the above-mentioned problems, we propose a panoramic probability map (PPM) guided detection system based on two cameras, where the proposed system can achieve high-speed object detection in a wide surveillance area.
Specifically, a wide-angle camera is used to provide prior information about the working scene based on the segmentation results of the captured panoramic image. Next, we use the prior information to guide the search camera in capturing detailed images of the targets. To achieve efficient target detection and accurate localization, we introduce two components: PPM and the probability searching module.
The PPM estimates high-probability regions where targets are likely to appear and provides approximate location estimation for some targets. 
Subsequently, the probability searching module utilizes the uncertainty variance obtained from the object detector to dynamically adjust the sampling range and refine the coordinates of the detected targets. This approach not only enhances the search efficiency but also improves the accuracy of target localization. 
Finally, by combining the PPM and probability searching modules, we develop a wide-area object detection system.
The main contributions of this research are as follows: 
\begin{itemize}
    \item We propose a novel PPM prior information generation method based on segmentation results from a wide-angle camera. The scanning process can focus on the generated high-probability regions to reduce cost.
    \item   We introduce a probability searching module based on the PPM to improve detection efficiency. The searching module can adjust the searching range by estimating the uncertainty of related objects, which shows adaptive searching ability in wide-area multiobject searching.
    \item  We implemented a hybrid wide-area multiobject detection system and conducted extensive indoor/outdoor experiments to verify the performance of the system. 
\end{itemize}

\section{Hybrid Wide-Area Multiobject Detection System with Probability Searching } \label{Third}

\begin{figure}[tpb]
	\begin{center}
		\includegraphics[width=0.95\columnwidth]{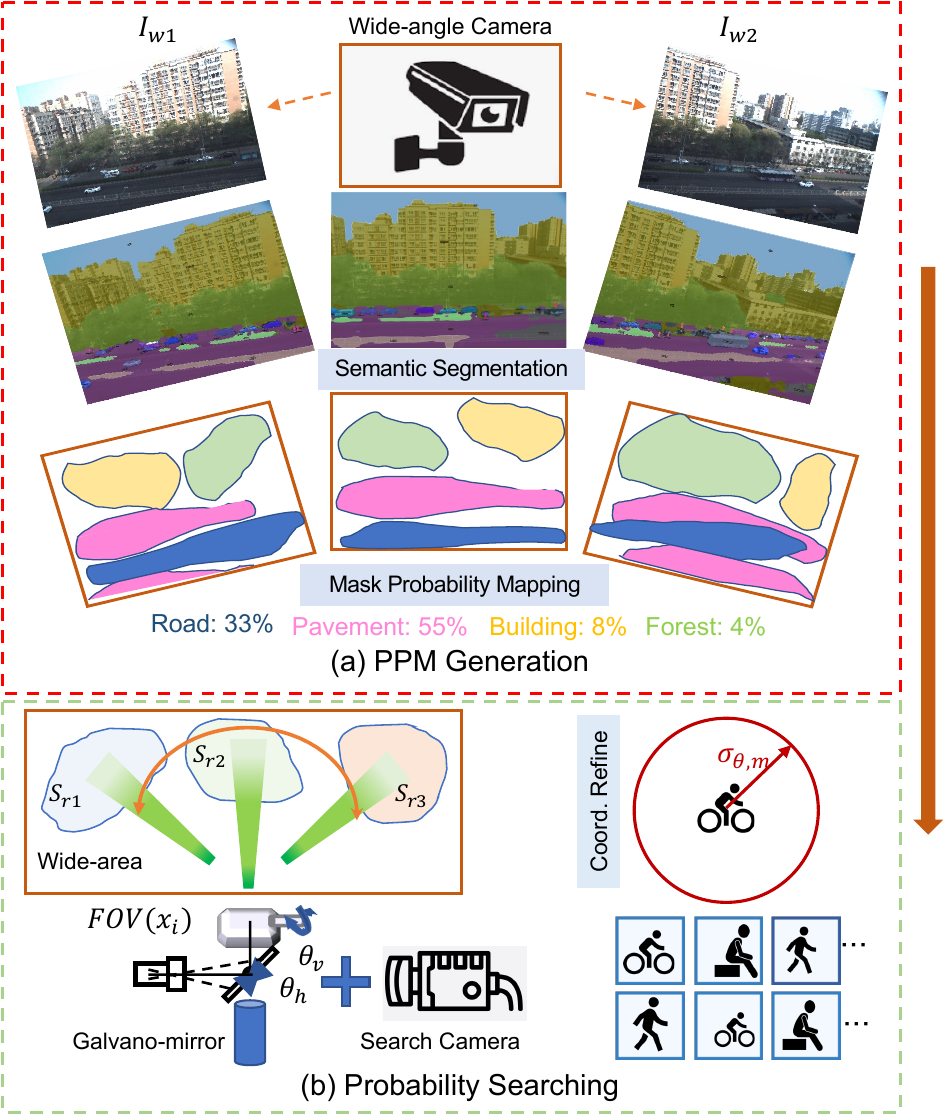}
		\caption{Illustration of the proposed hybrid wide-area multiobject detection system. (a) shows the generation process of PPM, which includes three steps: panoramic image $I_w$ capture, semantic segmentation, and mask probability mapping. (b) is the proposed probability searching method, it includes galvano-mirror searching and scanning.}

		\label{fig:system_ovw}
	\end{center}
\end{figure}

In Fig.\ref{fig:system_ovw}-(a), drawing insights from prior research\cite{RPM_guided_PF, Visual_Genome}, we utilize the segmentation outcomes to categorize each semantic region into distinct probability regions based on its label and spatial location, subsequently assigning varying probability values accordingly.
Following this, the particle sampling process is initiated, generating a set of particles primarily within high-probability regions conducive to target detection. To facilitate this, a galvano-mirror scanning module functionaly switch the search camera to capture the environmental information within the particle FOV. However, it is noteworthy that the initial object localization may exhibit deviations from the actual object position by several pixels. In response, a probability searching module assumes a crucial role in refining the object boundaries to rectify such errors.
As illustrated in Fig.~\ref{fig:system_ovw}-(b), we introduce a probability searching module in our proposed system. This module can dynamically adjust sampling range and refine the particle coordinates. These adjustments are made based on the uncertainty variance computed by the object detector, enhancing the precision and accuracy of object localization within our wide-area visual detection system.


\subsection{PPM Generation for Region Sampling} \label{sec:PPM_generation}


\subsubsection{{Semantic Segmentation-based PPM Generation}} \label{sec:PPIG}
As shown in Fig.~\ref{fig:system_ovw}-(a), PPM generation first performs semantic segmentation on the images captured by the wide-angle camera to obtain a segmentation map of the entire area. This serves as the foundation for assigning different sampling probabilities to different regions in the subsequent steps. This approach is based on the prior that the probability of targets appearing in different regions is different. 

We capture a panoramic image $I_w$ with the wide-angle camera. Then we fed this image into a semantic segmentation model (SemSegModel)~\cite{Ov_Seg} to obtain a segmentation map of the entire area:
\begin{equation}
\begin{gathered}
S_1 \cup S_2 \cup \cdots \cup S_n = \operatorname{SemSegModel}\left(I_w\right) \\
\qquad S_r \cap S_j=\phi, \forall r \neq j
\end{gathered}
\label{eq:seg}
\end{equation}
where $S_r, r=1,2,...,n$ represents the different regions after segmentation. Then, we assign a sampling probability $F(S_r,\mathcal{T})$ to each segmented region $S_r$.
\begin{equation}
\begin{gathered}
{F}\left(S_r, \mathcal{T}\right)=\frac{P\left(S_r\right) \times P\left(\mathcal{T} \mid S_r\right)}{\sum_{r=1}^n P\left(S_r\right) \times P\left(\mathcal{T} \mid S_r\right)}\\
P\left(S_r\right)=\frac{\text { Area }_{S_r}}{\text { Area }_{I_w}}, r=1,2, \ldots, n
\end{gathered}
\label{eq:F_Sr_O}
\end{equation}
where $P(S_r)$ represents the relative area of the current region $S_r$ to the entire image $I_w$. $P(\mathcal{T}|S_r)$ is the semantic probability of target-region, which is obtained by statistical analysis of object-region relationships in the Visual Genome dataset~\cite{Visual_Genome}. Through the steps above, the semantic segmentation model divides the search scene into several regions and determines the sampling probability $F(S_r,\mathcal{T})$ for each region $S_r$.  

Building upon panoramic area segmentation, we introduce panoramic object detection to generate the PPM. The PPM refines the sampling probabilities within each region, aiming to minimize searching in ineffective areas and enhance the accuracy of target localization. To achieve this, we use the PPM segmentation results to select `obejcts' as the initial results, to get $B_{\boldsymbol{O}}$,
where $o_i$ is the initial detected object, $O=(o_1,o_2,...,o_n)$ is the collection of detected objects during semantic segmentation in Eq.~\ref{eq:seg}, and $n$ is the number of segmentation objects. To simplify the estimation of uncertainty probability, we use the output of the segmentation results' confidence as the uncertainty $\sigma_O=B_{\boldsymbol{O}}$~\cite{Gaussian_YOLOv3}.

If a target $o$ is detected in region $S_r$, a new sampling region $S_{ro}$ is defined around the target $o$, and the corresponding sampling probability is denoted as $F(S_{ro},\mathcal{T})$. Assume the number of particles allocated to region $S_r$ is determined as $x_r=\mathcal{F}\left(S_r, \mathcal{T}\right) \times N$. The number of particles assigned to the region $S_{ro}$ is denoted as $x_{ro}$, while the remaining region $S_{rm}$ within $S_r$ receives $x_{rm} = x_r-x_{ro}$ uniformly distributed particles. The calculation for $x_{ro}$ is as follows:
\begin{equation}
\begin{gathered}
x_{ro}=\frac{F\left(S_{ro}, \mathcal{T}\right) \times S_{ro}}{F\left(S_{ro}, \mathcal{T}\right) \times S_{ro}+F\left(S_r, \mathcal{T}\right) \times S_{rm}} \cdot x_r\\
S_{ro}=\pi\left(r \sigma_o\right)^2 \\
S_{rm} = S_r - S_{ro} \\
\end{gathered}
\end{equation}
where $F(S_{ro},\mathcal{T})$ represents the sampling probability of the detected target $o$, which is set to 100\% in this paper. The hyperparameter $r$ is set to 50, and $\sigma_o$ denotes the uncertainty variance of object $o$ in Eq.~\ref{eq:seg}. 

Therefore, by dynamically adjusting the search area, the refined PPM optimizes the search process. It focuses on refining the localization of targets with small uncertainty while expanding the search for targets with high uncertainty.

\subsubsection{{Particle Sampling in High-Probability Regions}} \label{sec:PIS}  
\begin{figure}[tpb]
	\begin{center}
		\includegraphics[width=0.8\columnwidth]{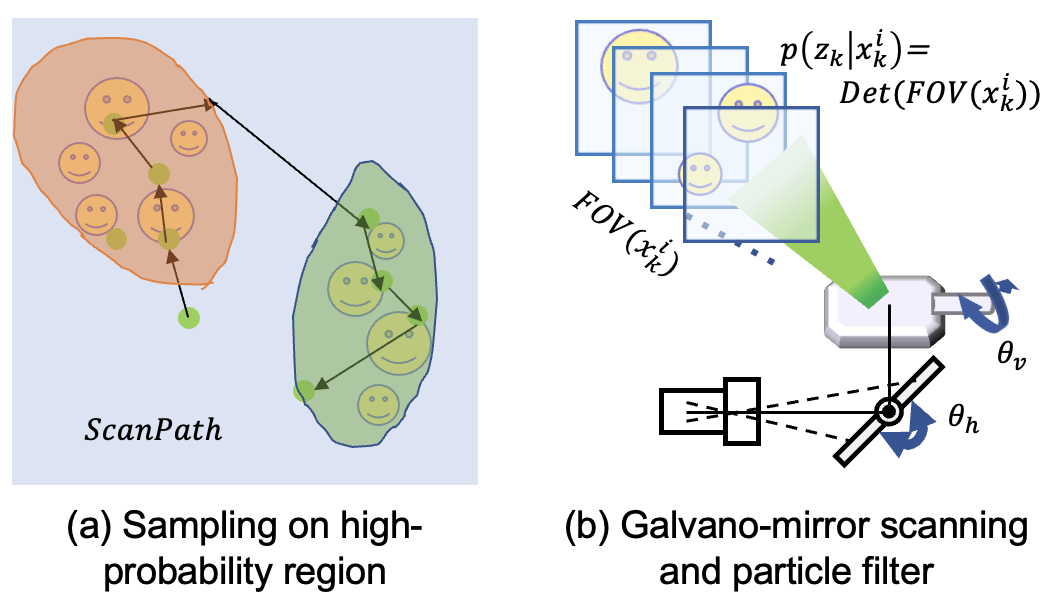}
		\caption{Visulization of high-probability region sampling and galvano-mirror scanning using particle filter.}
		\label{fig:sampling_scanning}
	\end{center}
\end{figure}

The particle sampling module searches for the target of interest by generating a set of particles $\boldsymbol{x}_k$. The entire particle sampling process is based on a pre-selected importance sampling proposal distribution function $q(\boldsymbol{x}_{k}^{i}|\boldsymbol{x}_{k-1}^{i},z)$:
\begin{equation}
\boldsymbol{x}_{k}^{i} \sim q(\boldsymbol{x}_{k}^{i}|\boldsymbol{x}_{k-1}^{i},z) , i=1,2,...,N_k,
\end{equation}	
where $N_k$ is the number of particles at stage $k$, $z$ is the whole wide-area FOV, and $\boldsymbol{x}_k^i$ is the $i^{th}$ particle at stage $k$.
The visualization of particle sampling based on the PPM high-probability regions is shown in Fig.~\ref{fig:sampling_scanning}-(a), where several particles (indicated by green circles) are sampled in segmentation regions.

During the normal iteration process, the proposal distribution function $q(\boldsymbol{x}_{k}^{i}|\boldsymbol{x}_{k-1}^{i},z)$ is updated based on the dynamically updated particles and their weights:
\begin{equation}
q(\boldsymbol{x}_{k}^{i}|\boldsymbol{x}_{k-1}^{i},z)=\sum_{i=1}^{\hat{N}_{k-1}}\tilde{w}_{k-1}^{i}*\mathcal{N}(\tilde{w}_{k-1}^{i},\sigma_{k-1}^2 I),
\label{eq:q_k}
\end{equation}
These parameters are obtained from the weight update module at time $k-1$. where $\tilde{w}_{k-1}^i$ is the normalized weight of particle $i$ (Eq.~\ref{eq:w_normilize}), $\hat{N}_{k-1}$ is the number of particles retained after removing redundant particles. $\sigma_{k-1}$ represents the uncertainty variance of object detection (Eq.~\ref{eq:object_detector}), which determines the sampling range for each particle. 

In the initial detection stage, we use the PPM to allocate sampling probabilities. As shown in the particle sampling part in Fig.~\ref{fig:sampling_scanning}-(a), we first assign sampling probabilities $q(\boldsymbol{x}_{0}^{i})=\mathcal{F}\left(S_r, \mathcal{T}\right)$ to each region based on the semantic map. Subsequently, we perform fine-grained allocation of sampling probabilities within each region, based on the detected targets and their corresponding uncertainties.

\subsubsection{{High-probability Region Scanning Based on PPM}} \label{sec:VmS}
By moving the galvano-mirror, it determines whether the target is within the FOV of the particle and updates the particle's weight accordingly. If particle $\boldsymbol{x}_k^i$ detects the object of interest, the probability of the target existing in the current visual image is calculated:
\begin{equation}
p\left(z \mid x_k^i\right),\sigma_k=\text {Detector}\left(\operatorname{Img}\left(\boldsymbol{x}_k^i\right)\right) \label{eq:object_detector}
\end{equation} 
$\sigma_k$ represents the uncertainty variance generated by the object detector, where hard samples such as small or occluded objects often exhibit large uncertainty. This process is shown in Fig.~\ref{fig:sampling_scanning}-(b), the system controls the galvano-mirror to scan the particles that lie in different high-probability regions.

The weight of each particle is updated based on the measurement model of the likelihood function, which is derived from the probability of detecting the object. The specific formula used for this calculation is:
\begin{equation}
w_{k}^{i}=w_{k-1}^{i} p(z|\boldsymbol{x}_{k}^{i}),i=1,2,...,N_k.
\label{eq:w_update}
\end{equation}
where the weight of each particle $w_{k}^{i}$, is determined by the previous weight and the likelihood probability $p(z|\boldsymbol{x}_{k}^{i})$. The likelihood probability is calculated by the object detection model (as shown in Eq.~\ref{eq:object_detector}).
\subsection{Probability Searching for Object's Coordinate Refinement} \label{sec:RPR}
\begin{figure}[t]
	\centering
	\includegraphics[width=0.7\linewidth]{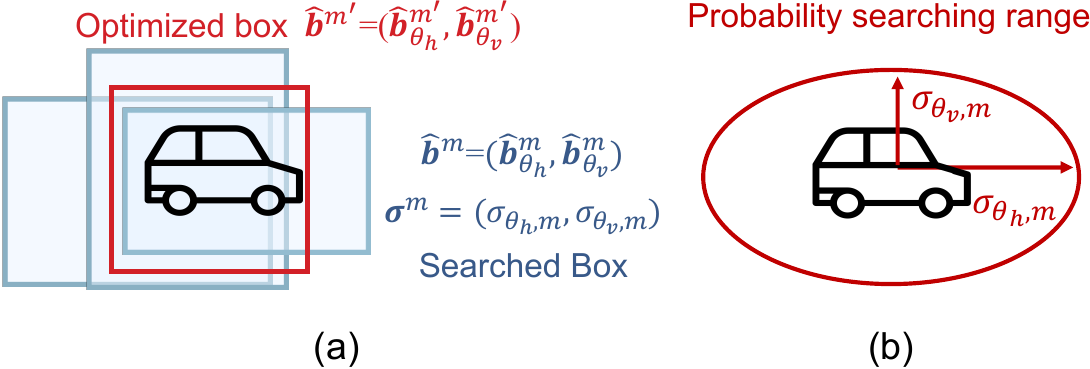}
	\caption{Coordinate refinement $m$ by the probability searching. (a) The red rectangle on the left is the object's boundary after adjustment. (b) The red circle on the right is the sampling range during probability searching.}
     \label{fig:variance_voting}
\end{figure}

As shown in Fig.~\ref{fig:variance_voting}, we propose to use probability searching to adjust the coordinates of the retained objects for more accurate positioning and enable efficient searching.

\subsubsection{Coordinate Refinement} \label{sec:coord refine}
The image coordinate system of particles is transformed into the galvano-mirror scanning coordinate system using the following equations: 
\begin{equation}
\left\{\begin{array}{l}
\hat{g}_{\theta_h}=\theta_h+\alpha \cdot \left(t_x-W / 2\right)  \\
\hat{g}_{\theta_v}=\theta_v+\alpha \cdot \left(t_y-H / 2\right) 
\end{array}\right.
\end{equation}
where $(\theta_h,\theta_v)$ is the coordinates of the particle $x_k^i$ in the galvano-mirror scanning system, $(t_x,t_y)$ is the coordinates of the target $\boldsymbol{b}$ in the image coordinate system, $(\hat{g}_{\theta_h},\hat{g}_{\theta_v})$ represent the refined coordinates of the searched object in the galvano-mirror system, $\alpha$ is a transform coefficient, which is 0.002 in this paper, and $(W/2, H /2)$ is the image center. 

Then, we perform NMS on the refined detection objects. If there are multiple nearby objects, only the one with the highest confidence score is retained; otherwise, both objects are kept. After performing NMS on all refinement searched objects $\hat{\boldsymbol{b}}$, the remaining search windows $\widehat{SW}_k$ are:

\begin{equation}
\widehat{SW}_k \cup \text{NMS}(\hat{\boldsymbol{b}}^i, \hat{\boldsymbol{b}}^j), \forall i, j \in\left[1,2, \ldots, N_k\right]
\end{equation}

\subsubsection{Probability Searching}
After adjusting the coordinates of the retained object, the search particle center is more accurate.
Initially, for each retained object $\hat{\boldsymbol{b}}^i$, we compute the coordinates and uncertainty variances $p_i$ of each object $\hat{\boldsymbol{b}}^i$ that overlap with it. 
\begin{equation}
    p_i =e^{-\left(1-\mathrm{IoU}\left(\hat{\boldsymbol{b}}^i, \hat{\boldsymbol{b}}^m\right)\right)^2 / \sigma_t}, \quad \forall i, \operatorname{IoU}\left(\hat{\boldsymbol{b}}^i, \hat{\boldsymbol{b}}^m\right)>0
\end{equation}

We then improve the coordinates of object $m$ through a voting process during NMS as follows:
\begin{equation}
\hat{b}_{\theta_h}^{m'}  =\frac{\sum_i p_i \hat{b}_{\theta_h}^i / \sigma_{\theta_h, i}^2}{\sum_i p_i / \sigma_{\theta_h, i}^2}, \quad \hat{b}_{\theta_v}^{m'}  =\frac{\sum_i p_i \hat{b}_{\theta_v}^i / \sigma_{\theta_v, i}^2}{\sum_i p_i / \sigma_{\theta_v, i}^2},\\
\label{eq:variance_voting}
\end{equation}
where $\hat{\boldsymbol{b}}^{m'}=\left(\hat{b}_{\theta_h}^{m'}, \hat{b}_{\theta_v}^{m'}\right)$ is the refined coordinates of object $m$, $\operatorname{IoU}\left(\hat{\boldsymbol{b}}^i, \hat{\boldsymbol{b}}^m\right)>0$ represents all objects that overlap with object $m$, $(\sigma_{\theta_h}, \sigma_{\theta_v})$ is the uncertainty variance in the $\theta_h$ and $\theta_v$ dimensions of object $i$, and $\sigma_t$ is a hyper-parameter set to 0.025 in this study.
As illustrated in Fi.~\ref{fig:variance_voting}, left red rectangle is the refined object boundary, $(\sigma_{\theta_h,m},\sigma_{\theta_v,m})$ is the radius of the adjusted sampling range. As shown in Fig.~\ref{fig:variance_voting}-(b), the following search procedure only needs to be processed in the optimized range. 
During the voting process, two types of adjacent objects receive lower weights: (1) objects with high uncertainty variance, and (2) objects with small IoU with the selected object. 


%
After removing potential duplicate sampled particles, we update the weight of each particle. 
All particles in $\widehat{SW}_k$ are weight-normalized as follows:
\begin{equation}
\widehat{w}_k^i=\frac{w_k^{\prime i}}{\sum_{i=1}^{\widehat{N}_k} w_k^{\prime i}}, i=1,2, \ldots, \widehat{N}_k
\label{eq:w_normilize}
\end{equation}
Based on the normalized weights, particles enter a new round of particle sampling to obtain a new iterative estimate of the target state (as shown in Eq.~\ref{eq:q_k}).





\section{Experiments} \label{5th}

Finally, the detection system is built up by combining the proposed PPM method and probability searching module. We conducted several experiments to verify its performance.


\subsection{Experimental Configuration} \label{sec:experimental_settings}
\begin{figure}[!t]
	\centering
        \includegraphics[width=0.9\linewidth]{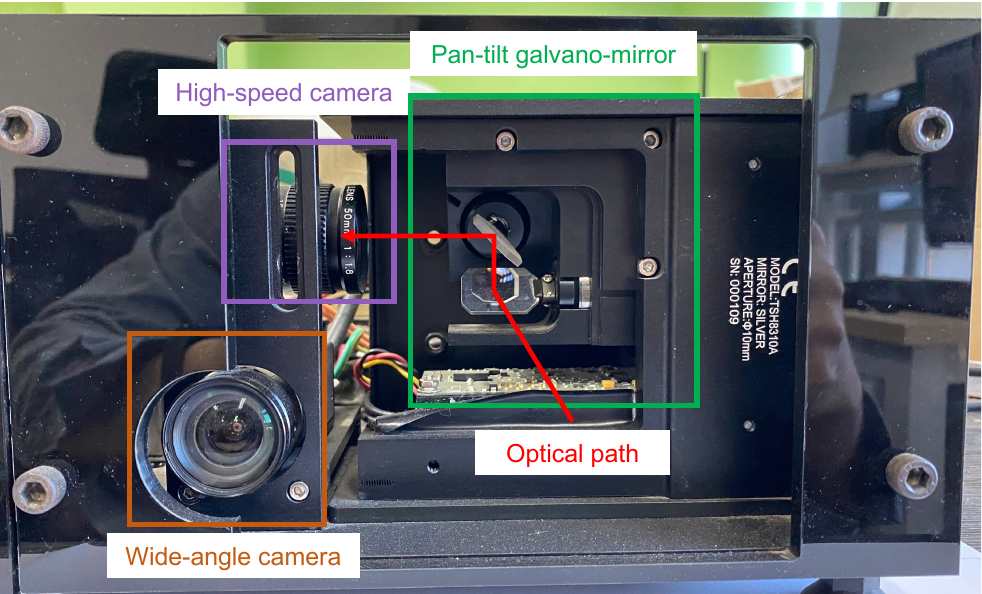}
	\caption{Construction of the system, which includes a wide-angle camera, a high-speed search camera, and a pan-tilt galvano-mirror.}
     \label{fig:sys_config}
\end{figure}

During our experiments, we employ a hybrid detection system by integrating a wide-angle camera (A5201CU150, Dahua, 1920$\times$1200 @ 150 fps) and a search camera (A7040CU000, Dahua, 640$\times$480 @ 437 fps). The focal length of wide-angle camera and search camera is 6 mm and 50 mm, respectively. To facilitate high-speed scanning, we utilize a dual-axis galvano-mirror (TSH8310A, Sunny Technology) to switch the FOV in a wide area. An overview of this system is presented in Fig.~\ref{fig:sys_config}. We control the whole system in a hardware-software integration manner by using a PC equipped with an Intel Core i7-8700K CPU @ 3.70 GHz, 16-GB DDR-4 memory, 256 GB of RAM, and a single Nvidia RTX 2080 Ti GPU. Cameras and the galvano-mirror are controlled by an FPGA control board. The image size of panoramic is 1440$\times$1200, while the search camera is 264$\times$224. The galvano-mirror's scanning FOV ranges from $-20^\circ$ to $20^\circ$, and its step response time is 0.25 ms.

\subsection{Object Detection Analysis of the Proposed System} \label{sec:object_detection}

\begin{figure}[t]
	\centering
	\includegraphics[width=0.9\linewidth]{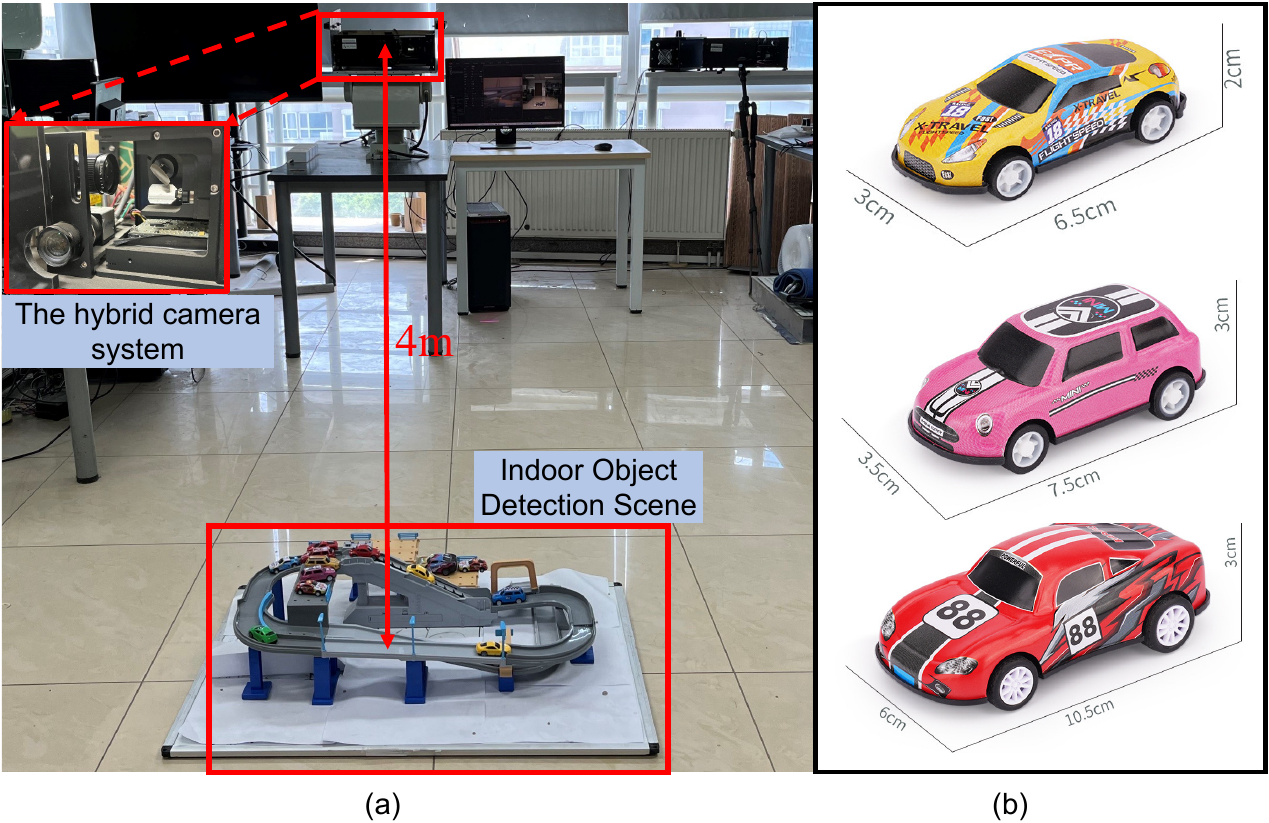}
	\caption{Indoor object detection setup for performance verification.}
     \label{fig:object_detection_config}
\end{figure}

To conduct a rigorous performance evaluation and ensure a fair comparison with the MPF, we conduct real-time object detection experiments in indoor environments. The experimental setup, as illustrated in Fig.~\ref{fig:object_detection_config}, involves several toy cars within a 4-meter radius of the detection system. The captured image size is 224$\times$224 pixels. In the captured images, the three types of toy cars are varying pixel dimensions: 94$\times$29, 101$\times$43, and 144$\times$43, respectively. These toy cars exhibit dynamic movements, including outlier toy cars, to comprehensively verify the system's detection performance.

\begin{figure}[t]
	\centering
	\includegraphics[width=\linewidth]{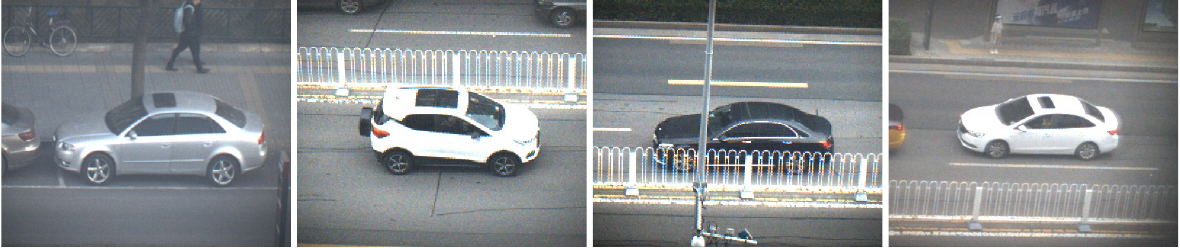}
	\caption{The multiobject detection results in a transportation monitoring scene.}
     \label{fig:outside_show}
\end{figure}

We conduct outdoor experiments to show the PPM-based wide-area detection system's performance. As shown in Fig.~\ref{fig:outside_show}, the detected objects lie in the center of the search camera, which indicates the probability searching method has object refinement ability and can search fast-moving objects clearly and efficiently.
The detection results of the MPF and our proposed algorithm for hard cases are shown in Fig.~\ref{fig:exp_1}. The hard cases in the experiment include large-sized vehicles that are difficult to capture in a single FOV, cars that are occluded, and cars that are outliers and have a side-facing pose. The detection results captured by the wide-angle camera enable our method to locate large-sized vehicles accurately. Moreover, due to the improved search efficiency brought by the PPM and the probability searching module, our method can locate small targets with occlusion and pose changes better.

\subsection{Recall Comparison of Different Searching Methods}
\begin{figure}[t]
	\centering
        \includegraphics[width=0.8\linewidth]{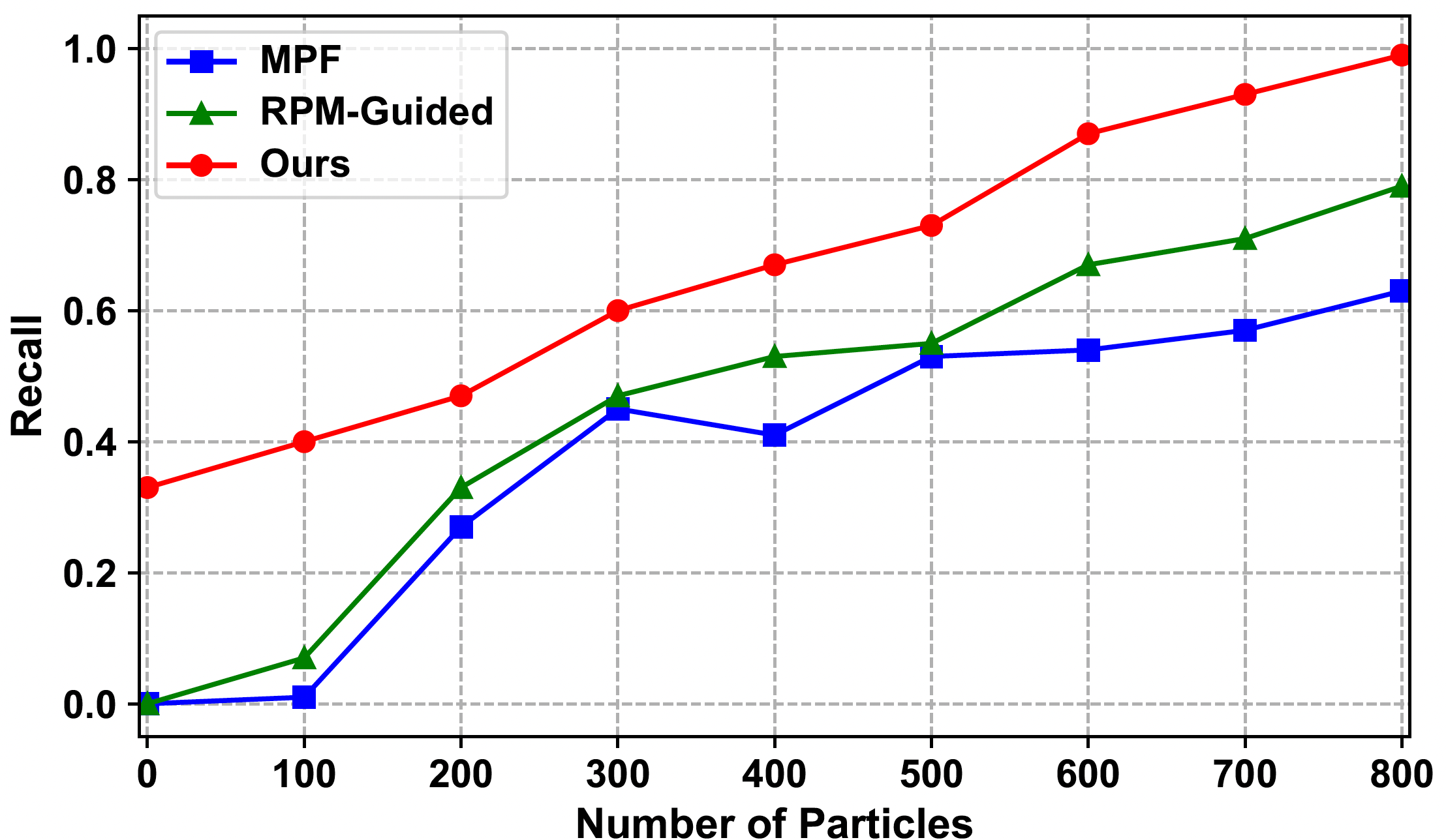}
	\caption{Recall comparison of different methods. We conduct separate evaluations of the MPF~\cite{Long}, RPM-guided PF~\cite{RPM_guided_PF}, and the final performance achieved by combining the PPM and probability searching modules (Ours).}
     \label{fig:exp_1}
\end{figure}

To validate the efficiency of our algorithm, we present our sampling process and compare it with MPF~\cite{Long} and RPM-guided PF~\cite{RPM_guided_PF} algorithms. Compared with these methods, our advantages mainly lie in the PPM and the probability searching modules. To verify their effectiveness, we deploy the PPM and the probability searching module to obtain the final result (Ours). As shown in Fig.~\ref{fig:exp_1}, by adding the probability searching module, our method can locate targets faster than MPF and RPM-guided PF, achieving a recall of 0.60 with 300 sampled particles and a recall of 0.67 with 400 sampled particles. The PPM can locate partial targets based on a wide-angle camera and obtain a region probability distribution map before sampling. Thus, before the searching process, our method has an initial detection result that is superior to other region sampling-based methods, e.g., MPF and RPM. 
As shown in Fig.~\ref{fig:exp_1}, the PPM curve reaches a recall of 0.33 at a particle number of 0, and by detecting difficult targets, its recall reaches 1.00 at a particle number of 800. By combining the PPM and probability searching, our method always reaches the highest recall rate at different particle settings. Fig.~\ref{fig:exp_1} indicates our approach outperforms all the comparison methods throughout the entire sampling process and can detect all objects in the scene when the number of particles is 800. In contrast, MPF needs to perform random scanning of the entire scene, resulting in low scanning efficiency and unstable results. For example, when the number of sampled particles increases from 300 to 400, the recall value of MPF decreases from 0.47 to 0.40.

\begin{table}
	\centering
	\small
	\centering
	\caption{Recall comparison of different methods under different proportions of high-probability region settings}
	\setlength{\tabcolsep}{6pt}
	\begin{tabular}{l|c|c|c|c|c}
		\toprule
		\diagbox [width=8em,trim=l] {Method}{Proportion} & 27\% & 35\% & 41\%  & 49\% & 63\% \\
		\hline
            Omni-Pano~\cite{Three_Axis} & 0.51 & 0.62 & 0.66 &  0.72 & 0.73 \\ \hline
		MPF~\cite{Long} & 0.63 & 0.72 & 0.76 & 0.82 &0.84\\ \hline
		RPM-Guided~\cite{RPM_guided_PF} & 0.89 & 0.87 & 0.84 & 0.83 & \textbf{0.85} \\ \hline
            PPM (Ours) & \textbf{0.93} & \textbf{0.90} & \textbf{0.87} & \textbf{0.86} & \textbf{0.85} \\
		\bottomrule
	\end{tabular}\vspace{0cm}
	\label{tab: recall_cmp}
\end{table}

To further demonstrate the superiority of the proposed PPM prior information with probability searching, we compare the recall rate of different multiobject searching methods under different proportions of high-probability regions. The results are listed in  Table~\ref{tab: recall_cmp}.
In this experiment, we compare the PPM method against three region probability-based algorithms, which include the Omni-Pano~\cite{Three_Axis}, MPF searching~\cite{Long}, and RPM-guided detection method~\cite{RPM_guided_PF}. During the experiment, we changed the proportion of high-probability regions from 27\% to 63\%. As can be seen from Table~\ref{tab: recall_cmp}, owing to the PPM prior information generation module, our method gets the best recall rates at different proportions. When the high-probability region is about 27\%, the searching particles are sampled in a compact region, resulting in dense searching, and the recall rate is the highest, reaching 0.93. When the high-probability region enlarges, the search performance is always better than other methods.

\subsection{Ablation Study of PPM and Probability Searching}
Since PPM gives the search process prior information, we conduct an ablation study to analyze the role of the PPM module.
We record the results when several object detection models are deployed in the system in Table~\ref{tab:ablation_study}. Specifically, we separately employ YOLOv3~\cite{Yolov3}, YOLOv8-M~\cite{yolov8_ultralytics} and YOLO-NAS-M~\cite{yolo_nas} as the CNN detector in Eq.~\ref{eq:object_detector}. Then, the ablation study is classified into a searching framework with and without PPM guidance. We statistic the recall rate, the AP value, and the processing speed, respectively. These three detection models suffer severe performance degradation when there is no PPM module (w/o). E.g., the proposed object searching method with YOLOv3 which contains PPM (w/) gets a 0.88 recall rate and 0.92 AP. However, when PPM is missed, the recall rate and AP decrease to 0.53 and 0.64, respectively. The best performance is the searching framework with YOLO-NAS-M, the recall rate reaches 0.94, and the AP is 0.96. 
It is noted that because of the panoramic segmentation process in PPM generation, there is a slight decrease in speed when the system employs the PPM method; e.g., YOLOv8-M decreases from 122.46 fps to 100.16 fps, and YOLO-NAS-M decreases from 133.27 fps to 120.98 fps.

Through the above ablation study, we can conclude that the PPM prior information generation is critical in the wide-area multiobject search framework, which generates high probability guidance for the following searching module. 

\begin{table}[t]
    \centering
    \small
    \centering
  \caption{Ablation study of the multiobject search framework without (w/0) and with (w/) PPM prior generation}
    \begin{tabular}{cccc}
    \toprule
    Object Detector & Recall & AP & speed (fps) \\
    \midrule
    YOLOv3~\cite{Yolov3}  (w/o)  & 0.53 &  0.64  & 88.18 \\
    YOLOv3~\cite{Yolov3}  (w/) & 0.88 &  0.92  & 79.37   \\
    \midrule
    YOLOv8-M~\cite{yolov8_ultralytics} (w/o)  &    0.60   & 0.75   & 122.46\\
    YOLOv8-M~\cite{yolov8_ultralytics} (w/)   &    0.93   & 0.95   & 100.16 \\
    \midrule
    YOLO-NAS-M~\cite{yolo_nas} (w/o)  &   0.64  & 0.79   & \textbf{133.27} \\
    YOLO-NAS-M~\cite{yolo_nas} (w)    &   \textbf{0.94}  & \textbf{0.96} &  120.98 \\
    \bottomrule
    \end{tabular}%
  \label{tab:ablation_study}%
\end{table}%

\begin{figure}[t]
	\centering
	\includegraphics[width=1.0\linewidth]{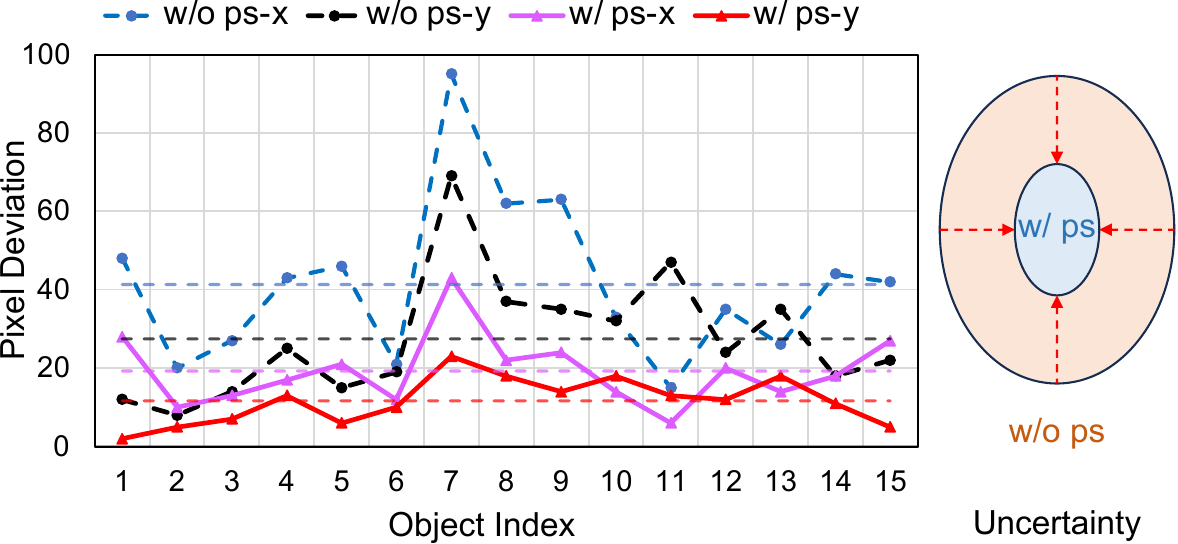}
	\caption{Ablation study of the probability searching module. The $x$ and $y$ are pixel deviations of each target from the image center after being calibrated by probability searching (represented as w/ ps-x, w/ ps-y) and without probability searching (represented as w/o ps-x, w/o ps-y). The dashed line is the mean of each axis. The right part shows the uncertainty before and after uncertainty searching, which gives less uncertainty after searching. 
 }
    \label{fig:pixel_dev}
\end{figure}

Then, we investigate the coordinate refinement performance of the probability searching module.
In Fig.~\ref{fig:pixel_dev}, the pixel deviations of each target from the image center are depicted to compare the sustained gaze effect on the target. We compare the performance of our method using probability searching (w/ ps-x, w/ ps-y) to refine the target boundary (Eq.~\ref{eq:variance_voting}) with that of not using the uncertainty module (w/o ps-x, w/o ps-y). As depicted in Fig.~\ref{fig:pixel_dev}, our method achieves smaller positioning deviations by utilizing probability searching. In particular, for moving targets (target 7, target 8, and target 9), probability searching can significantly reduce the positioning deviation. For target 7, without using the probability searching module, the positioning deviation in the x-axis and y-axis are 95 pixels and 69 pixels, respectively. By adding probability searching, the positioning deviation in the x-axis decreases to 43 pixels and the positioning deviation in the y-axis decreases to 23 pixels. 
Besides, the uncertainties are reduced after estimation and iterative searching, as shown in the right part of Fig.~\ref{fig:pixel_dev}. 
This experiment demonstrates the effectiveness of the proposed probability searching method.

\section{CONCLUSIONS} \label{6th}


 This paper introduces a novel wide-area hybrid vision system designed to address the challenges of multiobject detection. The system integrates a wide-angle camera and a search camera, thereby facilitating the capture of high-resolution images of objects and achieving precise object detection. To overcome the limitations of prior methods, we propose two pivotal components: the PPM prior information generation and the probability searching module. The former is dedicated to estimating regions of high probability for target appearance, significantly enhancing search efficiency. The latter dynamically adjusts the sampling range and refines target coordinates through uncertainty variances. The successful integration of these two modules culminates in the realization of a wide-area multiobject detection system. Comprehensive experiments demonstrate the system's exceptional capability to detect multiobject in large scenarios.






\bibliographystyle{IEEEtran}
\bibliography{root}

\end{document}